%% file: vimco_full.tex
\documentclass{article}
\usepackage{times}
\usepackage{graphicx} % more modern
\usepackage{subfigure}

\usepackage{natbib}

\usepackage{algorithm}
\usepackage{algorithmic}

\usepackage[T1]{fontenc}
\usepackage{amsmath,amsfonts,epsfig}
\usepackage{amssymb}

\usepackage{hyperref}

\newcommand{\pder}[1]{\nabla_{#1}}
\newcommand{\ddTheta}{\pder{\theta}}

\newcommand{\ddPsi}{\pder{\psi}}
\newcommand{\eqn}[1]{Eq.~\ref{#1}}
\newcommand{\sect}[1]{Section~\ref{#1}}
\newcommand{\tbl}[1]{Table~\ref{#1}}
\newcommand{\fig}[1]{Figure~\ref{#1}}
\newcommand{\wrt}{w.r.t.~}

\DeclareMathOperator*{\lse}{LogSumExp}
\DeclareMathOperator*{\softmax}{SoftMax}
\DeclareMathOperator*{\Sum}{Sum}

\newcommand{\varL}{{\mathcal L}}

\newcommand{\Ihat}{{\hat{I}}}
\newcommand{\Ihath}{{\hat{I}(h^{1:K})}}
\newcommand{\Lhat}{{\hat{L}}}
\newcommand{\Lhath}{{\hat{L}(h^{1:K})}}
\newcommand{\Lj}{{\hat{L}(h^j|h^{-j})}}
\newcommand{\Wtil}{{\tilde{w}}}
\newcommand{\EQ}{E_{Q(h^{1:K}|x)}}
\newcommand{\EQj}{E_{Q(h^j|x)}}
\newcommand{\EQmj}{E_{Q(h^{-j}|x)}}

\usepackage[accepted]{icml2016}

\begin{document} 

\twocolumn[
\icmltitle{Variational Inference for Monte Carlo Objectives}

\icmlauthor{Andriy Mnih}{amnih@google.com}
\icmlauthor{Danilo J. Rezende}{danilor@google.com}
\icmladdress{Google DeepMind}

\icmlkeywords{variational inference, generative models, importance sampling}

\vskip 0.3in
]

\begin{abstract}
Recent progress in deep latent variable models has largely been driven by the
development of flexible and scalable variational inference methods. Variational
training of this type involves maximizing a lower bound on the log-likelihood,
using samples from the variational posterior to compute the required gradients.
Recently, \citet{burda2015importance} have derived a tighter lower
bound using a multi-sample importance sampling estimate of the likelihood and
showed that optimizing it yields models that use more of their capacity
and achieve higher likelihoods.  This development showed the importance of such
multi-sample objectives and explained the success of several related
approaches.

We extend the multi-sample approach to discrete latent variables and analyze
the difficulty encountered when estimating the gradients involved. We then
develop the first unbiased gradient estimator designed for importance-sampled
objectives and evaluate it at training generative and structured output
prediction models.  The resulting estimator, which is based on low-variance
per-sample learning signals, is both simpler and more effective than the NVIL
estimator \citep{mnih2014neural} proposed for the single-sample variational
objective, and is competitive with the currently used biased estimators.
\end{abstract} 

\section{Introduction}

Directed latent variable models parameterized using neural networks have
recently enjoyed a surge in popularity due to the recent advances in
variational inference methods that made it possible to train such models
efficiently. These methods
\citep{kingma2013auto,rezende2014stochastic,mnih2014neural} approximate the
intractable posterior of the model with a variational posterior parameterized
using a neural network and maximize a lower bound on the intractable marginal
log-likelihood, estimating the required gradients using samples from the
variational posterior. This approach implements an efficient feedforward
approximation to the expensive iterative process required by traditional
variational inference methods for each data point.

One important weakness of variational methods is that training a powerful
model using an insufficiently expressive variational posterior can cause the model
to use only a small fraction of its capacity. The most direct route to
addressing this issue is to develop more expressive but still tractable
variational posteriors as was done in
\citep{salimans2015markov,rezende2015variational, gregor2015draw}.

However, the crippling effect of an excessively simple posterior on the model
can alternatively be seen as a consequence of the form of the lower bound
optimized by the variational methods \cite{burda2015importance}. As the bound
is based on a single-sample estimate of the marginal likelihood of the
observation, it heavily penalizes samples that explain the observation poorly
and thus produce low estimates of the likelihood. As result, the variational
posterior learns to cover only the high-probability areas of the true
posterior, which in turn assumes a simpler shape which is easier to
approximate by the variational posterior. A simple way to minimize this effect
is to average over multiple samples when computing the marginal likelihood
estimate. The resulting lower bound on the log-likelihood gets tighter as the
number of samples increases \cite{burda2015importance}, converging to the true
value in the limit of infinitely many samples. We will refer to such
objectives derived from likelihood estimates computed by averaging over
independent samples as \textit{Monte Carlo objectives}. When using
an objective that averages over multiple samples, the distribution for
generating samples no longer explicitly represents the variational posterior
and instead is thought of as a \textit{proposal distribution} due to
connections to importance sampling.

Multi-sample objectives of this type have been used for generative modelling
\citep{bornschein2015reweighted,burda2015importance}, structured output
prediction \citep{raiko2014techniques}, and models with hard
attention \cite{ba2015learning}. As a multi-sample objective is a better proxy
for the log-likelihood than a single-sample one, models trained using
multi-sample objectives are likely to achieve better log-likelihoods.
This has been empirically demonstrated in the context of generative models by
\citet{burda2015importance} and \citet{bornschein2015reweighted}, who also
showed that using more samples in the objective increased the number of latent
variables used in the deeper layers.

Unfortunately, unless all the latent variables in the model are continuous,
learning the proposal distribution with a multi-sample objective is difficult
as the gradient estimator obtained by differentiating the objective has very
high variance. As a result, with the exception of
\citet{burda2015importance}, who used an alternative estimator available for
continuous latent variables, none of the above methods update the parameters of
the proposal distribution by following the gradient of the multi-sample
objective. Thus, updates for the proposal distribution and the model parameters
in these methods are not optimizing the same objective function, which can lead
to suboptimal performance and even prevent convergence.

In this paper we develop a new unbiased gradient estimator for multi-sample
objectives that replaces the single learning signal of the naive estimator
with much lower variance per-sample learning signals.  Unlike the NVIL
estimator \citep{mnih2014neural} designed for single-sample variational objectives, our
estimator does not require learning any additional parameters for variance
reduction. We expect that the availability of an effective unbiased gradient
estimator will make it easier to integrate models with discrete latent
variables into larger systems that can be trained end-to-end.

\section{Multi-sample stochastic lower bounds}

\subsection{Estimating the likelihood}
\label{sec:est_like}

Suppose we would like to fit an intractable latent variable model $P(x, h)$
to data.  As the intractability of inference rules out using maximum likelihood
estimation, we will proceed by maximizing a lower bound on the log-likelihood.
One general way to derive such a lower bound is to start with an unbiased
estimator $\Ihat$ of the marginal likelihood $P(x)$ and then
transform it.
We will consider estimators 
of the form $\Ihath = \frac{1}{K} \sum_{i=1}^K f(x, h^i)$ where $h^1, ...,
h^K$ are independent samples from some distribution $Q(h|x)$
which can potentially depend on the observation $x$.
Before showing how to transform such an estimator into a bound,
let us consider some possible choices for the likelihood estimator.

Perhaps the simplest estimator of this form can be constructed by sampling $h^i$'s
from the prior $P(h)$ and averaging the resulting conditional likelihoods:
\begin{align}   \label{eqn:prior_estimator}
    \Ihath  = \frac{1}{K} \sum\nolimits_{i=1}^K P(x|h^i)\text{ with }h^i \sim P(h).
\end{align}
While this estimator is unbiased, it can have very high variance in models
where most latent configurations do not explain a given observation well. For
such models, the estimator will greatly underestimate the likelihood for most
sets of $K$ independent samples and substantially overestimate it for a small
number of such sets. This is a consequence of not taking into account the
observation we would like the latent variables to explain when sampling them.

We can incorporate the information about the observation we are estimating
the likelihood for by sampling the latents from a \textit{proposal}
distribution $Q(h|x)$ conditional on the observation $x$ and using importance
sampling:
\begin{align}   \label{eqn:is_estimator}
    \Ihath = \frac{1}{K} \sum\nolimits_{i=1}^K \frac{P(x,h^i)}{Q(h^i|x)}
\end{align}
with $h^{1:K} \sim Q(h^{1:K}|x) \equiv \prod_{i=1}^{K} Q(h^i|x)$.
In addition to also being unbiased, the variance of this estimator can be much lower
than that of the preceding one because it can assign high probability to the
latent configurations with high joint probability with the given observation.
In fact, if we were able to use the true posterior as the proposal
distribution, the estimator would have zero variance. While this is
infeasible for the models we are considering, this fact suggests that making
the proposal distribution close to the posterior is a sensible strategy.

\subsection{Lower-bounding the log-likelihood}

Having chosen an estimator $\Ihat$ for the likelihood, we can obtain an
estimator $\Lhat$ of a lower bound on the log-likelihood simply by taking the
logarithm of $\Ihat$.
We can justify this by applying Jensen's inequality:
\begin{align*}
     \EQ \left[ \log \Ihath \right] & \le \log \EQ \left[ \Ihath \right] \\
                                    & =   \log P(x),
\end{align*}
where the equality follows from the fact that since $\Ihat$ is unbiased, $\EQ[\Ihath] = P(x)$. 
Therefore, we can think of $\Lhath=\log \Ihath$ as a stochastic lower bound on
the log-likelihood \citep{burda2015importance}.

We note that this approach is not specific to the to estimators from
\sect{sec:est_like} and can be used with any unbiased likelihood estimator
based on random sampling. Thus it might be possible to obtain better lower
bounds by using methods from the importance sampling literature
such as control variates and adaptive importance sampling.

Despite the potential pitfalls described above, estimators involving sampling
from the prior have been used successfully for training models for structured
output prediction \citep{tang2013learning,dauphin2015predicting} and models
with hard attention \citep{mnih2014recurrent,zaremba2015reinforcement}.

The multi-sample ($K>1$) version of the above estimator has recently been used for
variational training of latent variable models \citep{burda2015importance,
bornschein2015reweighted} as well as models with hard attention
\citep{ba2015learning}. The single-sample version of the estimator yields the
classical variational lower bound \citep{jordan99variational}
\begin{align}
    \label{eqn:var_objective}
    \varL(x) = E_{Q(h|x)} \left[ \log \frac{P(x,h)}{Q(h|x)} \right],
\end{align}
which is used as the objective in much of the
recent work on training generative models
\citep{kingma2013auto,rezende2014stochastic,mnih2014neural}.

The advantage of using a multi-sample stochastic lower bound is that
increasing the number of samples $K$ is guaranteed to make the bound tighter
\citep{burda2015importance}, thus making it a better proxy for the
log-likelihood.  Intuitively, averaging over the $K$ samples inside the $\log$,
removes the burden of every sample having to explain the observation well,
which leads to the proposal distribution being considerably less concentrated
than the variational posterior, which is its single-sample counterpart.
Training models by optimizing a multi-sample objective can be seen as a
generalization of variational training that does not explicitly represent the
variational posterior.

\subsection{Objective}

Thus we will be interested in training models by maximizing objectives of the form
\begin{align}
    \label{eqn:objective}
    \varL^K(x) %= & \EQ \left[ \log \Ihath \right] \nonumber \\
               = & \EQ \left[ \log \frac{1}{K} \sum\nolimits_{i=1}^K f(x, h^i) \right],
\end{align}
which can be seen as lower bounds on the log-likelihood. This class of objectives
is a rich one, including the ones used in variational inference,
generative modelling, structured prediction, and hard attention.

\subsection{Gradient analysis}
\label{sec:gradient_analysis}

In this section we will analyze the gradient of the objective \wrt the
parameters of the model and the proposal distribution and explain
why developing an effective unbiased estimator for the gradient
is difficult in general.
In the special case of continuous latent variables an alternative approach to
gradient estimation based on reparameterization \citep{kingma2013auto,
burda2015importance} is likely to be preferable to the more general approach we
follow in this paper, which is applicable to all types of latent variables.

As shown in the supplementary material,
differentiating $\varL^K(x)$ \wrt the parameters $\theta$ of $Q$ and $f$ gives
\begin{align}
    \label{eqn:proposal_grad}
    \ddTheta \varL^K(x) % = & \ddTheta \EQ \left[ \Lhath \right]  \nonumber \\
                        = & \EQ \left[ \sum\nolimits_j \Lhath \ddTheta \log Q(h^j|x) \right] + \nonumber \\
                          & \EQ \left[ \sum\nolimits_j \Wtil^j \ddTheta \log f(x, h^j) \right],
\end{align}
where  $\Wtil^j \equiv \frac{f(x, h^j)}{\sum_{i=1}^K f(x, h^i)}$.

As our objective $\varL^K(x)$ is an expectation of the stochastic lower bound
$\Lhath$ \wrt to the proposal distribution, it can depend on any given
parameter through the proposal distribution, through the value of the
stochastic lower bound as a function of a set of $K$ samples, or both.
Intuitively, the first and the second terms in \eqn{eqn:proposal_grad} capture
the effect of $\theta$ on $\varL^K(x)$ though its effect on the proposal
distribution and the value of the stochastic lower bound as a function of a
set of samples respectively.

Let us inspect these two terms, both of which are linear combinations of the
gradients corresponding to the $K$ samples.  The second term is well-behaved
and is easy to estimate because weights $\{\Wtil^j\}$ are non-negative and
sum to 1, ensuring that the norm of the linear combination of the gradients is
at most as large as the norm of the largest of the $K$ gradients. In mixture
modelling terms, we can think of $\Wtil^j$ as the responsibility of sample $j$
for the observation $x$ --- a measure of how well sample $j$ explains the
observation compared to the other $K-1$ samples.

The first term however is considerably more problematic for two reasons. First,
the gradients for all $K$ samples are multiplied by the same scalar $\Lhath$, which 
can be thought of as the \textit{learning signal} for the proposal distribution
\citep{mnih2014neural}. As a result, the gradient for a sample that explains the
observation well is not given any more weight than the gradient for a sample in
the same set of $K$ that explains the observation poorly. This means that the
first term does not implement credit assignment \textit{within} each set of $K$
samples, unlike the second term which achieves that by weighting the gradients
using the responsibilities. Thus the learning signal for each sample
$h_i$ will have high variance, making learning slow.\footnote{ Despite not
performing credit assignment within sets of $K$ samples, the first term does
perform correct credit assignment in expectation over such sets.
}

Another important source of variance when estimating the first term is
the magnitude of the learning signal. Unlike the responsibilities used in the
second term, which are between 0 and 1, the learning signal can have potentially
unbounded magnitude, which means that the norm of the first term can become
much larger than the norm of any of the individual sample gradients. This issue
can be especially pronounced early in training, when all samples from the
proposal $Q$ explain the data poorly, resulting in a very small $\Ihath$ and
thus a very negative learning signal. Thus unless special measures are taken,
the first term in the gradient will overwhelm the second term and make the
overall estimate very noisy.

\subsection{Gradient estimation}

The difficulties described in the previous section affect only the gradient for
the parameters the sampling distribution depends on. For all other parameters
$\psi$ the first term is identically zero, which leaves only the second,
well-behaved term. As a result, the following naive Monte Carlo estimator based
on a single set of $K$ samples works well:
\begin{align} \label{eqn:model_grad}
    \ddPsi \varL^K(x) \simeq  \textstyle \sum_j \Wtil^{j} \ddPsi \log f(x, h^{j}),
\end{align}
where $h^{i} \sim Q(h|x)$.
While it is possible to reduce the variance of the estimator by averaging over
multiple sets of samples, in this paper we follow the common practice of using
a single set and relying on averaging over the training cases in a minibatch to
reduce the variance to a reasonable level instead.
We will now turn out attention to the more challenging problem of
estimating gradients for parameters that affect the proposal distribution.

\subsubsection{Naive}

We will start with the simplest estimator, also based on naive Monte Carlo:
\begin{align}
    \ddTheta \varL^K(x) \simeq &\textstyle \sum_j \Lhath \ddTheta \log Q(h^j|x) \nonumber \\
                             + &\textstyle \sum_j \Wtil^j \ddTheta \log f(x, h^j),
\end{align}
with $h^i\sim Q(h|x)$. This estimator does not attempt to eliminate either of
the two sources of variance described in \sect{sec:gradient_analysis} and we include it here 
for completeness only.

\subsubsection{With baselines (NVIL)}
\label{sec:nvil}

One simple way to reduce the variance due the large magnitude of the learning
signal is to reduce its magnitude by subtracting a quantity, called a
\textit{baseline}, correlated with the learning signal but not dependent on the
latent variables. This transformation of the learning signal leaves the
gradient estimator unbiased because it amounts to subtracting a term which
has the expectation of 0 under the proposal distribution. In our use of baselines, we
will follow the Neural Variational Inference and Learning
\citep[NVIL,][]{mnih2014neural} method for training generative models, which
is based on optimizing the classical variational lower bound
(\eqn{eqn:var_objective}). The main idea behind the NVIL estimator is to reduce
the magnitude of the learning signal for the parameters of the variational
distribution (which is the single-sample counterpart of our proposal
distribution) by subtracting two baselines from it: a constant baseline $b$ 
and an input-dependent one $b(x)$.

The following estimator is a straightforward adaptation of the same idea to
multi-sample objectives:
\begin{align} \label{eqn:nvil}
    \ddTheta \varL^K(x) \simeq & \textstyle\sum_j (\Lhath - b(x) - b) \ddTheta \log Q(h^j|x) \nonumber \\
                             + & \textstyle\sum_j \Wtil^j \ddTheta \log f(x, h^j),
\end{align}
with $h^i\sim Q(h|x)$.
The constant baseline $b$ tracks the mean of the learning signal, while the
input-dependent one is fit to minimize the squared residual of the learning signal
$\Lhath-b(x)-b$, with the goal of capturing the effect of the observation on
the magnitude of the learning signal. We implement the input dependent baseline
using a one-hidden layer neural network.

While introducing baselines can addresses the estimator variance due to the
large magnitude of the learning signal, it has no effect on the variance resulting from 
having the same learning signal for all samples in a set of $K$.

\subsubsection{Per-sample learning signals}
\label{sec:local_signals}

We can reduce the effect of the second source of variance by defining a
different \textit{local} learning signal for each sample in a way that
minimizes its dependence on the other samples in the set. This can be accomplished
by using a separate baseline for each sample that depends on the value of all
other samples and eliminates much of the variance due to them. We will now show
that this approach does not bias the resulting estimator.

Let $h^{-j}$ denote the set of $K-1$ samples obtained by leaving out sample $j$
from the original set. Since the samples in a set are independent, evaluating
the expectations with respect to them in any order produces the same result.
Thus the contribution of sample $j$ to the first term in \eqn{eqn:proposal_grad} can be
expressed as
\begin{align*}
    \EQ \left[ \Lhath \ddTheta \log Q(h^j|x) \right] = \nonumber \\
       \EQmj \left[ \EQj \left[ \Lhath \ddTheta \log Q(h^j|x) \middle| h^{-j}\right] \right].
\end{align*}
Since in the inner-most expectation all samples except for $h^j$ are
conditioned on, adding any function of them to the learning signal for $h^j$
has no effect on the value of the expectation. Thus we can define a baseline
that depends on $h^{-j}$ in addition to $x$. We would like this baseline to be as
close to $\Lhath$ as possible without using the value of $h^j$.

Inspecting the global learning signal $\Lhath$ (\eqn{eqn:objective})
suggests that we can obtain an effective baseline for the learning signal for sample $j$
by replacing $f(x, h^j)$ in it by some quantity close to it
but independent of $h^j$. We could, for example, use some mapping
$f(x)$ trained to predict $f(x, h^i)$ from the observation $x$.
This gives rise to the following local learning signal for sample $j$:
\begin{align}
    \Lj = \Lhath - \log \frac{1}{K} \left( \sum\nolimits_{i \ne j} f(x, h^i) + f(x) \right). \nonumber
\end{align}
For $K=1$, this estimator becomes essentially equivalent to the NVIL estimator,
with $\log f(x)$ corresponding to the input-dependent baseline $b(x)$.

We can avoid having to learn an additional mapping by taking advantage of the
fact that we have more than one sample in a set. Since the samples in a set are
IID, so are the corresponding values $f(x, h^i)$, which means that we can get a
reasonable estimate $\hat{f}(x, h^{-j})$ by combining the $f(x,h^i)$ values for all
the other samples in the set using averaging of some sort. We experimented
with using the arithmetic mean
($\hat{f}(x, h^{-j}) = \frac{1}{K-1}\sum_{i \ne j} f(x, h^i)$) and
the geometric mean ($\hat{f}(x, h^{-j}) = \exp \left(\frac{1}{K-1}\sum_{i \ne j} \log f(x, h^i)\right)$)
and found that the geometric mean worked slightly better.
The resulting local learning signals can be written as
\begin{align}
    \label{eqn:vimco_estimator}
    \hat{L}(h^j|& h^{-j}) = \\
        & \Lhath - \log \frac{1}{K} \left (\sum\nolimits_{i \ne j}f(x, h^i) + \hat{f}(x, h^{-j})\right). \nonumber
\end{align}
This approach to variance reduction, unlike the one above or NVIL,
does not require learning any additional
parameters for performing variance reduction. Moreover, as the total cost of
computing the per-sample learning signals in \eqn{eqn:vimco_estimator} is of the
same order as that of computing of the global learning signal, this approach
allows us to implement effective variance reduction in the multi-sample case
essentially at no cost. This approach relies on having more than one sample for
the same observation, however, and so is not applicable in the single-sample
setting.

The final estimator has the form
\begin{align}
    \ddTheta \varL^K(x) \simeq & \sum\nolimits_j \Lj \ddTheta \log Q(h^j|x) \nonumber \\
                             + & \sum\nolimits_j \Wtil^j \ddTheta \log f(x, h^j).
\end{align}
We will refer to this estimator as the VIMCO (Variational Inference for Monte
Carlo Objectives) estimator.
The pseudocode for computing it is provided in the supplementary material.
This estimator is a black-box one, in the sense that it
can be easily applied to any model for which we can compute the complete
log-likelihood $\log P(x, h)$ and its parameter gradients exactly. As such, it
can be seen as as an alternative to Black Box Variational Inference
\citep{ranganath2014black} and NVIL, specialized for multi-sample objectives.

\section{Structured output prediction}

Structured output prediction (SOP) is a type of supervised learning with
high-dimensional outputs with rich structure such as images or text.  The
particular emphasis of SOP is on capturing the dependencies between the output
variables in addition to capturing their dependence on the inputs.
Here we will take the approach of viewing SOP as conditional probabilistic
modelling with latent variables \citep{tang2013learning,sohn2015learning}.

To stay consistent with the terminology for generative models we used so far,
we will refer to inputs as \textit{contexts} and to outputs as
\textit{observations} Thus, given a set of context/observation pairs $(c,x)$,
we would like to fit a latent variable model $P(x, h|c)$ to capture the
dependencies between the contexts and the observations, as well as those between
the observed dimensions. Typically such a model factorizes as
$P(x,h|c)=P(x|h,c)P(h|c)$, with both the conditional likelihood and the prior
terms being conditional on the context.  Thus, this is essentially the same
setting as for generative modelling, with the only difference being that every
distribution now also conditions on the context $c$, which makes it
straightforward to apply the estimators we presented.

However, historically such models have been trained using samples
from the prior $P(h|c)$, with the gradients computed using either importance
sampling \citep{tang2013learning} or heuristic rules for backpropagating
through binary units \citep{raiko2014techniques}. Since using the prior as the
proposal distribution does not allow it to use the information about the
observation, such methods tend to require a large number of samples to perform
well. Though variational training has been applied recently to SOP models with
continuous latent variables \citep{sohn2015learning}, we are not aware of any
work that uses a learned proposal distribution conditional on the observations
to train SOP models with multi-sample objectives. We will explore the
effectiveness of using this approach in \sect{sec:results_struct}.

\section{Related work}

\textbf{Multi-sample objectives:} The idea of using a multi-sample objective for
latent variable models was proposed by \citet{raiko2014techniques}, who
thought of it not as a lower bound on the log-likelihood but an objective in
its own right. They evaluated several gradient estimators at optimizing it for
training structured prediction models and showed that a simple biased estimator
emulating backpropagation performed best. \citet{tang2013learning} proposed an
estimator based on importance sampling for an EM-like bound on the
log-likelihood using samples from the prior. This is also a biased estimator as
it relies on self-normalized importance sampling to approximate the posterior
using a set of weighted samples.
\citet{burda2015importance} pointed out that the multi-sample objective of
\citet{raiko2014techniques} was a tighter lower bound on the log-likelihood
than the single-sample variational lower bound and presented a method for
training variational autoencoders by optimizing this multi-sample objective.
Their method relies on an unbiased gradient estimator which can be used only
for models with continuous latent variables.

\textbf{Reweighted Wake Sleep:}
Though the Reweighted Wake Sleep algorithm
\citep[RWS,][]{bornschein2015reweighted} for training generative models has
been derived from the perspective of approximating the log-likelihood gradients
using importance sampling, it is closely related to the bound optimization
approach we follow in this paper. \citet{burda2015importance} have shown that the RWS gradient
estimator for the model parameters is identical to the one given by
\eqn{eqn:model_grad}, which means that the RWS model parameter update aims
to maximize the lower bound on the log-likelihood based on the multi-sample
importance sampling estimator from \eqn{eqn:is_estimator}. RWS performs two
types of updates for the proposal distribution parameters, the first of which,
called the \textit{wake} update, is based on the same weights $\{\Wtil^j\}$ as
the model parameter update
\begin{align} \label{eqn:wake_grad}
    \Delta \theta \propto \textstyle \sum_j \Wtil^j \ddTheta \log Q(h^j|x)
\end{align}
and is motivated as a (biased) estimator of the gradient $KL(P(h|x)||Q(h|x))$.
Its bias decreases with the increasing number of samples, vanishing in the
limit of infinitely many samples.

The second update, called the \textit{sleep} update, having the form
\begin{align} \label{eqn:sleep_grad}
    \Delta \theta \propto \ddTheta \log Q(h|x),
\end{align}
is based on a sample $(x, h)$ from the model and
comes from the original Wake-Sleep algorithm \citep{HintonWakeSleep}.
The wake update tends to work better than the sleep update, and using the two
updates together works even better \citep{bornschein2015reweighted}.
As neither of these updates appears to be related to the lower bound optimized
the model parameter update, RWS does not seem to optimize a well-defined
objective, a feature it shares with the original Wake-Sleep algorithm. Despite
this theoretical weakness RWS works well in practice, outperforming 
original Wake-Sleep and NVIL, which are single-sample algorithms, using as
few as 5 samples per observation.

\textbf{Black Box Methods:}
As our approach does not assume anything about the structure of the model or
the distribution(s) of its latent variables, it can be seen as a black box
method for multi-sample objectives. A number of black box methods have been
developed for the classical variational objective, usually based around
unbiased gradient estimators for the proposal distribution. Black Box
Variational Inference \citep[BBVI][]{ranganath2014black} and NVIL
\citep{mnih2014neural} are two such methods.

VIMCO shares some similarities with the black box method of the local
expectations (LE) of \citet{titsias2015local}. The LE method provides a
relatively low variance unbiased estimator based on local learning signals
derived from computing an exact expectation \wrt each variable in the model.
Both methods work well without baselines and involve considering multiple
values for latent variables.  Unlike VIMCO, the LE method optimizes a
single-sample objective and requires computing exact expectations for each
variable, which makes it much more computationally expensive.

\begin{figure*}
\begin{minipage}{0.5\textwidth}
\includegraphics[width=.9\textwidth]{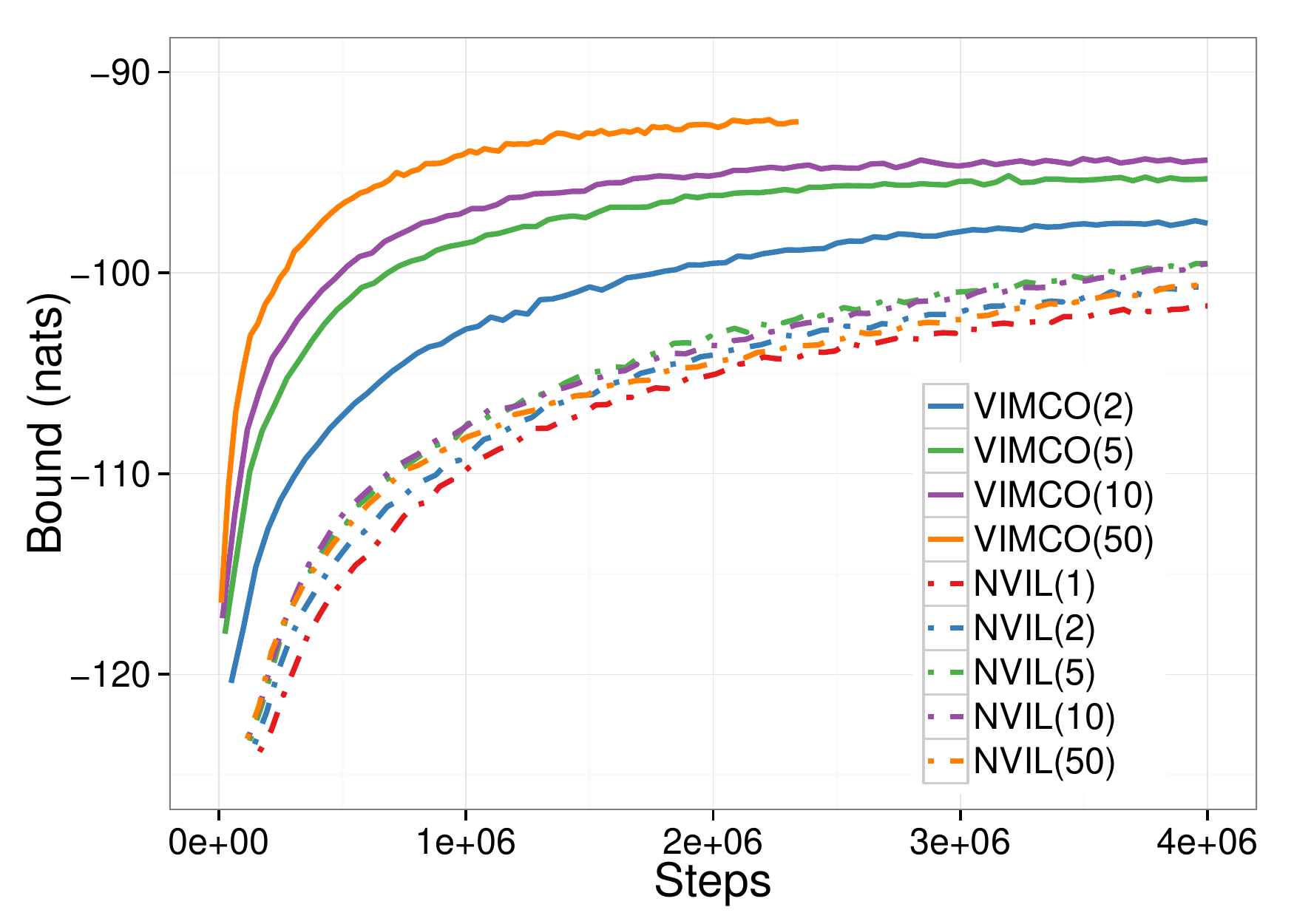}
\end{minipage}
\begin{minipage}{0.5\textwidth}
\includegraphics[width=.9\textwidth]{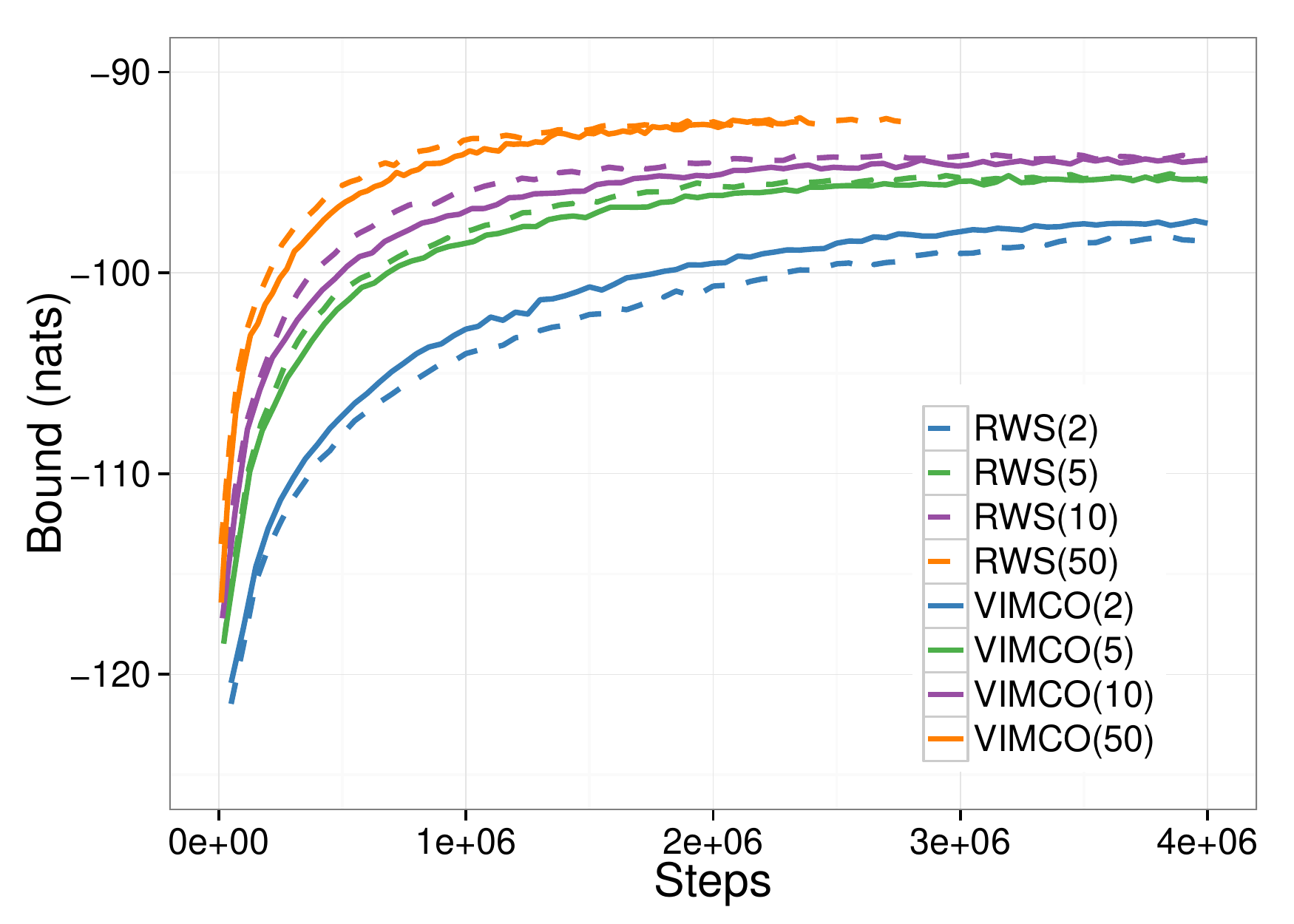}
\end{minipage}
\caption{Generative modelling: Comparison of multi-sample objective on the validation set for the SBNs
trained on MNIST using VIMCO and those trained using (Left) NVIL and (Right)
Reweighted Wake Sleep. The number in brackets specifies the number of samples
used in the training objective.}
\label{fig:mnist_bounds}
\end{figure*}

\section{Results}

We evaluate the effectiveness of the proposed approach at training models for
generative modelling and structured output prediction. We chose these two tasks
because they involve models with hundreds of latent variables, which poses
formidable challenges when estimating the gradients for the proposal
distributions.
In both cases we compare the performance of the VIMCO estimator to that of the
NVIL estimator as well as to an effective biased estimator from the literature.
We experiment with varying the number of samples in the objective to see how
that affects the performance of the resulting models when using different
estimators. The details of the training procedure are given in the supplementary
material.

\subsection{Generative modelling}
\label{sec:results_generative}

We start by applying the proposed estimator to training generative models,
concentrating on sigmoid belief networks (SBN)
\citep{neal1992connectionist} which consist of layers of binary latent
variables. SBNs have been used to evaluate a number of variational training
methods for models with discrete latent variables
\citep{mnih2014neural,bornschein2015reweighted,gu2015muprop}.

Our first comparison is on the MNIST dataset of $28\times28$ images of
handwritten digits, using the binarization of \citet{salakhutdinov2008} and
the standard 50000/10000/10000 split into the training, validation, and
test sets. We use an SBN with three hidden layers of 200 binary latent
variables (200-200-200-768) as the generative model. The proposal distribution is
parameterized as an SBN with the same architecture but going in the opposite
direction, from the observation to the deepest hidden layer (768-200-200-200).

As our primary goal is here is to see how well the VIMCO estimator performs at
optimizing the multisample objective, we train the above model using each of
the VIMCO, NVIL, and RWS estimators to optimize the lower bound
(\eqn{eqn:objective}) based on 2, 5, 10, and 50 samples ($K$).
To match the computational complexity of the other two estimators, we used only
the better-performing wake update for the proposal
distribution in RWS.
We also trained the model by optimizing the
classical variational objective ($K=1$) using NVIL to serve as a single-sample baseline.
In all cases, the model
parameter gradients were estimated using \eqn{eqn:model_grad}.

\fig{fig:mnist_bounds} shows the evolution of the training objective on the
validation set as training proceeds. From the left plot, which compares the models trained
using VIMCO to those trained NVIL, it is apparent that VIMCO is far more
effective than NVIL at optimizing the multi-sample objective and benefits much
more from using more samples. NVIL performance improves slightly when using a
modest number of samples before starting to degrade upon reaching $K=10$. The right plot shows the
comparison between VIMCO and RWS. The two methods perform similarly, with VIMCO
performing better when using 2 samples and RWS learning slightly faster when using more samples.

Having selected the best model for each method/number of samples combination
based on its validation score, we estimated its negative log-likelihood on
the test set using 1000 proposal samples for each data point. The results
in \tbl{tbl:gen_mnist} show that VIMCO and NVIL perform slightly
better than RWS for 2 samples. However, as the number of samples
increases, VIMCO and RWS performance steadily improves while
NVIL performance stays virtually the same until reaching $K=50$, when it becomes
markedly worse. Overall, RWS and VIMCO perform similarly,
though VIMCO seems to have a slight edge over RWS for all
numbers of samples we considered.

We also investigated the effectiveness of VIMCO and NVIL variance reduction
techniques more directly, by monitoring the magnitude of their learning signals
during training. While VIMCO and NVIL performed comparably when using the
2-sample objective, VIMCO benefited much more from using more samples.
For the 10-sample objective, the average magnitude of the VIMCO learning signal
was 3 times lower than that of NVIL.  More details are given in the
supplementary material.

\begin{figure*}
\begin{minipage}{0.5\textwidth}
\includegraphics[width=.9\textwidth]{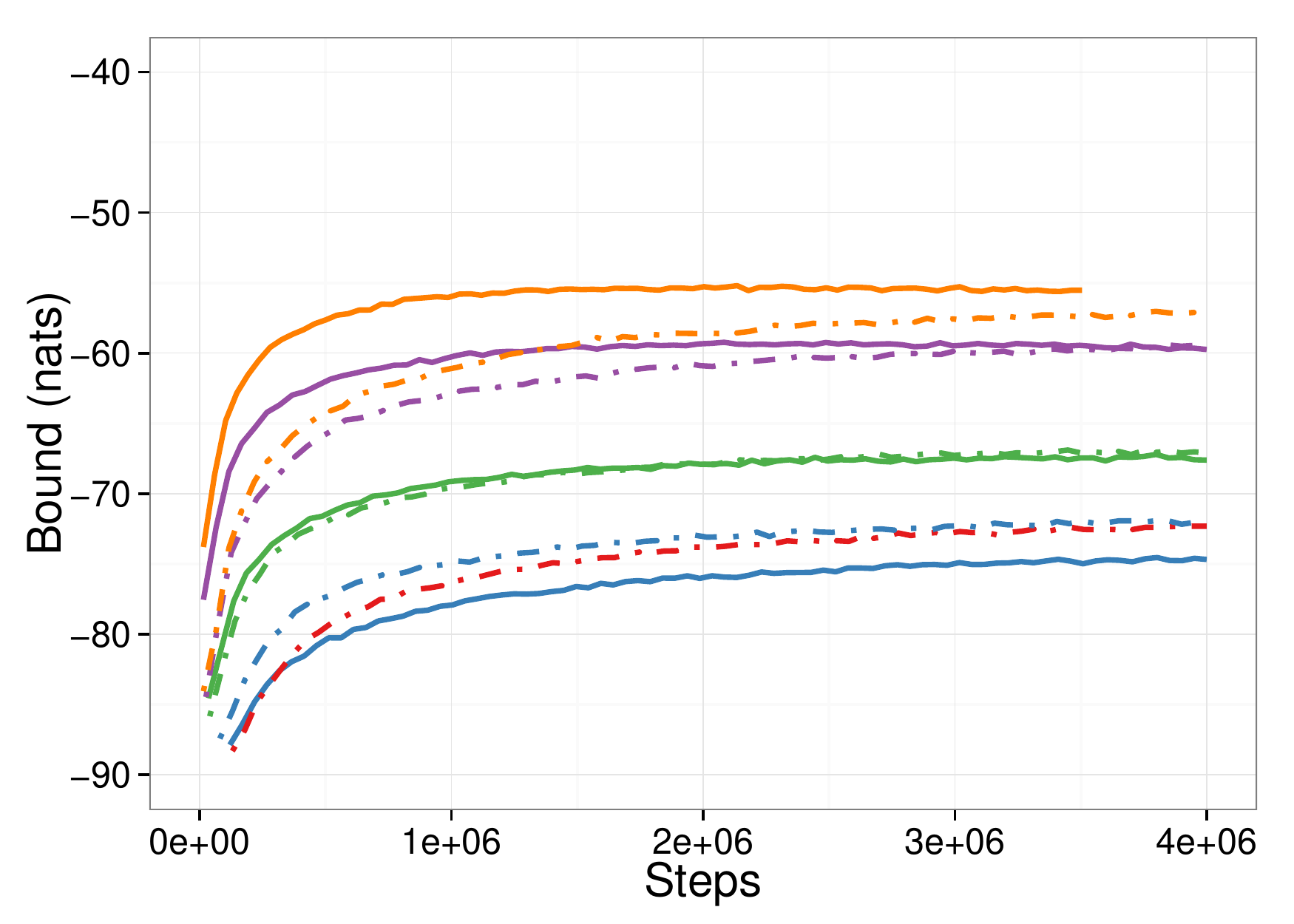}
\end{minipage}
\begin{minipage}{0.5\textwidth}
\includegraphics[width=.9\textwidth]{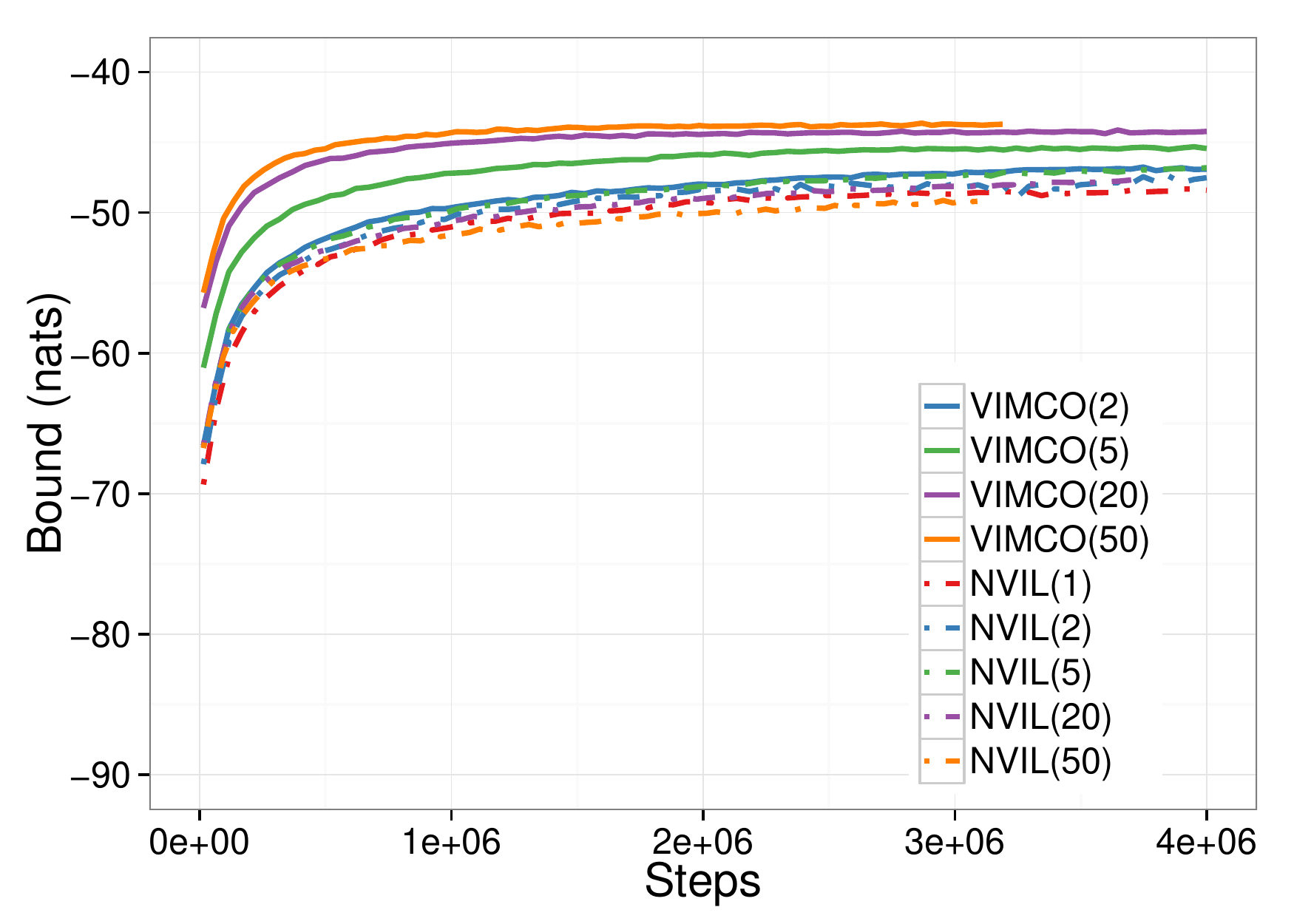}
\end{minipage}
\caption{Structured output prediction: Comparison of multi-sample objective on
the validation set for a 3-hidden-layer SBN trained with VIMCO against those trained
with NVIL using sampling from (Left) the prior and (Right) the learned
proposal distribution. The number in brackets specifies the number of samples
used in the training objective.}
\label{fig:cond_mnist_bounds}
\end{figure*}

\begin{table}[t]
\caption{Estimates of the negative log-likelihood (in nats) for generative modelling on MNIST.
The model is an SBN with three latent layers of 200 binary units.}
\label{tbl:gen_mnist}
\begin{center}
\begin{small}
\begin{sc}
\begin{tabular}{|c|c|c|c|}
\hline
Number  of         &               \multicolumn{3}{|c|}{Training alg.}  \\
samples            &     \multicolumn{1}{|c|}{VIMCO} & \multicolumn{1}{|c|}{NVIL} &  \multicolumn{1}{|c|}{RWS}  \\
\hline
1                  &   ---         &   95.2   &  ---   \\
2                  &   93.5        &   93.6   &  94.6  \\
5                  &   92.8        &   93.7   &  93.4  \\
10                 &   92.6        &   93.4   &  93.0  \\
50                 &   91.9        &   96.2   &  92.5  \\
\hline
\end{tabular}
\end{sc}
\end{small}
\vspace{-0.35in}
\end{center}
\end{table}

\subsection{Structured output prediction}
\label{sec:results_struct}

In the second set of experiments we evaluated the proposed estimator at
training structured output prediction models. We chose a task that has been
used as a benchmark for evaluating gradient estimators for models with binary
latent variables by \citet{raiko2014techniques} and \citet{gu2015muprop}, which
involves predicting the lower half of an MNIST digit from its top half.  We
trained two SBN models, one with two and one with three layers of 200 binary
latent variables between the 392-dimensional ($14\times28$) input and output
layers. We use the same binarized MNIST dataset for this task as for the
generative modelling experiments in \sect{sec:results_generative}.

We consider two different kinds of proposal distributions for training
the models. In the first case, we follow the standard practice for training
structured output prediction models and use the model prior as the proposal
distribution. However, as the prior does not have access to the observation
information which is available during training, most of the resulting samples
are unlikely to explain the observation well, potentially leading to
inefficient use of samples and unnecessarily noisy learning signal. Hence, in
the second case we learn a separate proposal distribution that takes 
both the context and the observation halves of the image as input. We parameterize the
proposal distribution using an SBN with the same structure as the prior except
that the last layer of latent variables in addition to being conditioned on the
preceding layer is also conditioned on the observation.

We train the models with VIMCO
and NVIL using 2, 5, 20, and 50 sample objectives. As in the previous experiment,
we also train single-sample baseline models using both
types of proposals with NVIL (and $K=1$). \fig{fig:cond_mnist_bounds} shows the
resulting multi-sample bound values for the three-layer models on the
validation set as a function of the number of parameter updates. The left plot,
containing the results for models trained by sampling from the prior, shows
that model performance improves dramatically as the number of samples is
increased. Though NVIL with 1 or 2 samples, performs better than VIMCO with 2
samples, as the number of samples increases their roles reverse, with VIMCO
making much faster progress than NVIL for 20 and 50 samples. The fact that
increasing the number of samples has such an effect on model performance strongly
suggests that samples generated from the prior rarely explain the observation
well.

The right plot on \fig{fig:cond_mnist_bounds} shows the result of training
with a learned proposal distribution. It is clear that using a learned
proposal leads to drastic improvement for all method / number of samples
combinations. In fact, the worst result obtained using a learned proposal
distribution is better than the best result obtained by sampling from the
prior. In terms of relative performance, the story here is similar to that from
the generative modelling experiment: VIMCO performs better than NVIL and
benefits much more from increasing the number of samples. The gap between the
methods is considerably smaller here, likely due to the task being
easier. Inspecting the conditional digit completions 
sampled from the models shows that the models trained using a learned proposal
distribution capture multimodality inherent in the task very well. We show
conditional completions from a three-layer model trained using VIMCO with 20
samples in the supplementary material.

Finally, to compare to the results of \citet{raiko2014techniques}, we followed
their evaluation protocol and estimated the negative log-likelihoods for the
trained models using 100 samples. Their best result on this task
was 53.8 nats, obtained using a 2-layer SBN trained using a biased estimator
emulating backprop to optimize the 20-sample objective.  With VIMCO training,
the same model achieves 56.5 nats using the prior as the proposal and 46.1 nats
with a learned proposal, which is the first sub-50 nat result on this task.

\section{Discussion}

In this paper we introduced VIMCO, the first unbiased general gradient
estimator designed specifically for multi-sample objectives that generalize the
classical variational lower bound. By taking advantage of the structure of the
objective function, it implements simple and effective variance reduction at no
extra computational cost, eliminating the need for the learned baselines
relied on by other general unbiased estimators such as NVIL.

We demonstrated the effectiveness of VIMCO by applying it to variational
training of generative and structured output prediction models. It
consistently outperformed NVIL and was competitive with the currently used biased
estimators.

While classical variational methods can perform poorly when using an
insufficiently expressive variational posterior, multi-sample objectives
provide a graceful way of trading computation for quality of fit simply by
increasing the number of samples used inside the objective. Combining such
objectives with black box variational inference methods could
make the latter substantially more effective.
We thus hope that the proposed approach will increase the appeal and
applicability of black box variational inference.

\subsubsection*{Acknowledgements}
We thank Alex Graves, Guillaume Desjardins, Koray Kavukcuoglu, Volodymyr Mnih,
Hugo Larochelle, and M\'{e}lanie Rey for their helpful comments.

\renewcommand{\bibsection}{\section*{References}}

\bibliography{vimco}
\bibliographystyle{icml2016}

\newpage
\include{supplementary_include}

\end{document}

%% file: supplementary_include.tex
\section*{A. Algorithm for computing VIMCO gradients}

Algorithm~\ref{alg:vimco} provides an outline of our implementation of VIMCO
gradient computation for a single training case. This version uses the
geometric mean to estimate $f(x, h^j)$ from the other $K-1$ terms. The
computations are performed in the log domain for better numerical stability.

\begin{algorithm}[th!]
\caption{Compute gradient estimates for the model and proposal distribution parameters for a single observation}
\label{alg:vimco}
\begin{algorithmic}
    \REQUIRE $x$ , $K \ge 2$
    \FOR {$i = 1$ to $K$} 
        \STATE $h^i \sim Q(h|x)$ \\
        \STATE $l[i] = \log f(x, h^i) $ \\
    \ENDFOR  \\
    \COMMENT{Compute the multi-sample stochastic bound}
    \STATE $\Lhat = \lse(l) - \log K$ \\
    \COMMENT{Precompute the sum of $\log f$}
    \STATE $s = \Sum(l)$ \\
    \COMMENT{Compute the baseline for each sample} \\
    \FOR {$i = 1$ to $K$} 
        \STATE \COMMENT{Save the current $\log f$ for future use and replace it}
        \STATE \COMMENT{with the average of the other K-1 $\log f$ terms}
        \STATE $temp = l[i]$
        \STATE $l[i] = (s - l[i])/(K-1)$
        \STATE $\Lhat^{-i}  = \lse(l) - \log K$ \\
        \STATE $l[i] = temp$ \COMMENT{Restore the saved value}
    \ENDFOR 
    \STATE $w = \softmax(l)$ \COMMENT{Compute the importance weights} \\
    \STATE $\nabla\theta = 0, \nabla \psi = 0$
    \STATE \COMMENT{Sum the gradient contributions from the K samples}
    \FOR{$i = 1$ to $K$} 
        \STATE \COMMENT{Proposal distribution gradient contributions} \\
        \STATE $\nabla\theta = \nabla\theta + (\Lhat-\Lhat^{-i}) \ddTheta \log Q(h^i|x)$
        \STATE $\nabla\theta = \nabla\theta + w[i] \ddTheta \log f(x, h^i)$ 
        \STATE \COMMENT{Model gradient contribution} \\
        \STATE $\nabla\psi = \nabla\psi + w[i] \ddPsi \log f(x, h^i)$
    \ENDFOR  
\end{algorithmic}
\end{algorithm}

%\newpage
\section*{B. Details of the experimental protocol}

All models were trained using the Adam optimizer \citep{kingma2015adam} with
minibatches of size 24. The input to the proposal distribution/inference
network was centered by subtracting the mean.
For each training method/number of samples combination
we trained the model several times using different learning rates,
saving the model with the best validation score achieved during each training
run. The plots and the scores shown in the paper were obtained from the saved
model with the highest validation score. For generative training, we considered the
learning rates of $\{3\times 10^{-4}, 1\times 10^{-3}, 3\times 10^{-3}\}$.
For the structured output prediction experiments, the
learning rates were $\{3\times 10^{-4}, 1\times 10^{-3}, 3\times 10^{-3}\}$ for VIMCO and RWS
and $\{1\times 10^{-4}, 3\times 10^{-4}, 1\times 10^{-3}\}$ for NVIL.

Our NVIL implementation used both constant and input-dependent baselines as
well as variance normalization.  The input-dependent baseline for NVIL was a
neural network with one hidden layer of 100 tanh units. VIMCO used the
geometric mean for computing the per-sample learning signals.

\begin{figure}
\begin{minipage}{0.5\textwidth}
\includegraphics[width=.9\textwidth]{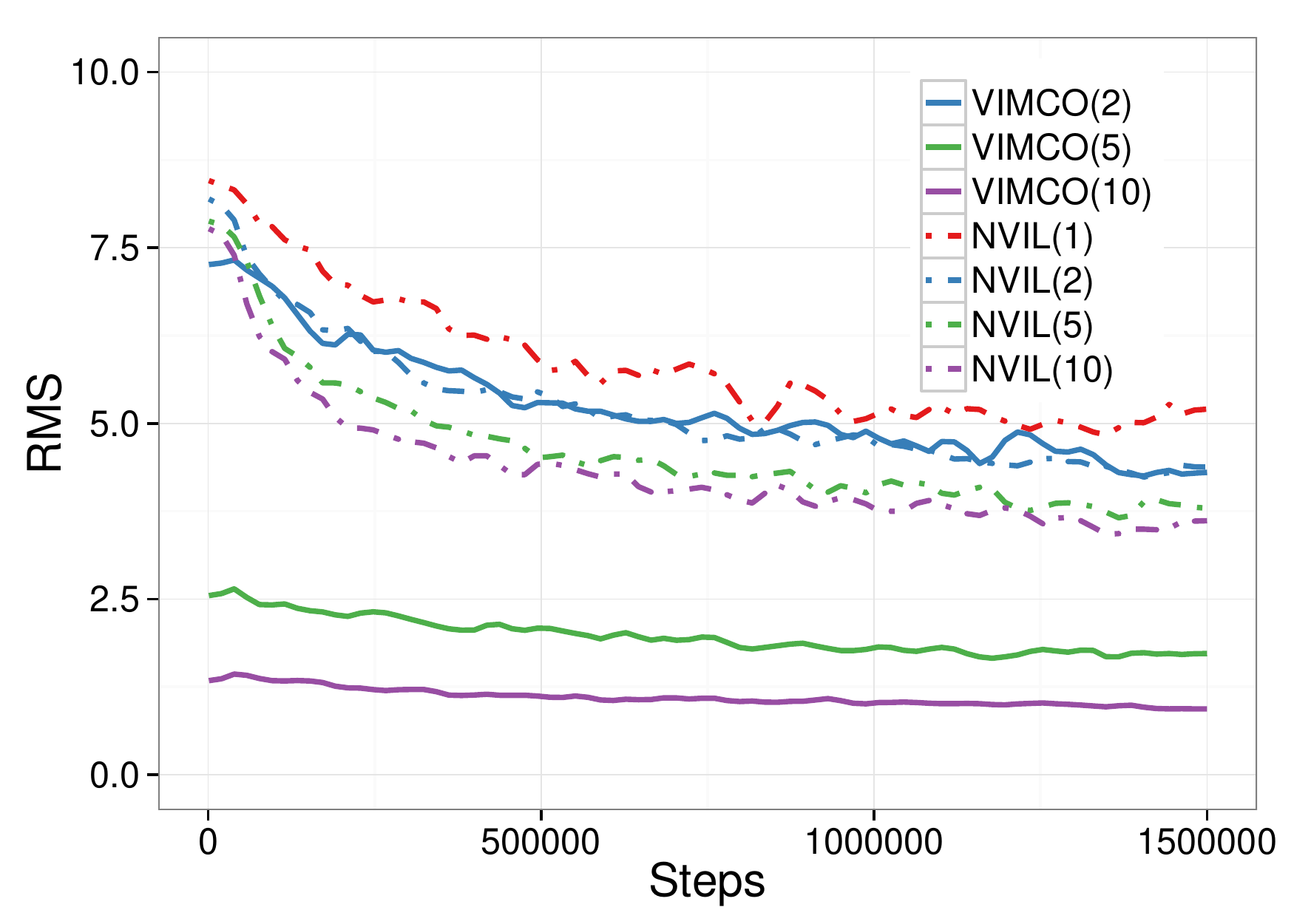}
\end{minipage}
\caption{
The magnitude (root mean square) of the learning signal for VIMCO and NVIL as a function of the
number of samples used in the objective and the number of parameter updates.
}
\label{fig:rms}
\end{figure}

\section*{C. Effect of variance reduction on the learning signal}

As explained in Sections~2.4 and 2.5, the magnitude of the learning signal used
for learning the proposal distribution parameters is closely related to the
variance of the resulting gradient estimator. Both VIMCO and NVIL aim to reduce
the estimator variance by subtracting a baseline from the original learning
signal $\Lhath$ in order to reduce its magnitude, while keeping the estimator
unbiased. We examined the effectiveness of these two approaches by plotting 
a smoothed estimate of the magnitude of the resulting learning signal ($\Lj$
for VIMCO and $\Lhath-b(x)-b$ for NVIL) as a function of the number of
parameter updates when training the SBN on MNIST in Section~5.1. The magnitude
of the learning signal was estimated by taking the square root of the mean of
the squared signal values for each minibatch. The results for different numbers of samples
shown in \fig{fig:rms} suggest that while VIMCO and NVIL are equally
effective at reducing variance when using a 2-sample objective, VIMCO becomes
much more effective than NVIL when using more than 2 samples. For 10 samples,
the average magnitude of the learning signal for VIMCO is about 3 times lower
than for NVIL, which suggests almost an order of magnitude lower variance
of the gradient estimates.

\section*{D. Gradient derivation for the multi-sample objective}

In this section we will derive the gradient for the multi-sample objective
\begin{align}
    \varL^K(x) = & \EQ \left[ \Lhath \right] \nonumber \\
               = & \EQ \left[ \log \Ihath \right] \nonumber \\
               = & \EQ \left[ \log \frac{1}{K} \sum_{i=1}^K f(x, h^i) \right]. \nonumber
\end{align}

We start by using the product rule:
\begin{align}
  \label{eqn:gradVarL}
    \ddTheta \varL^K(x) = & \ddTheta \EQ \left[ \Lhath \right]   \nonumber \\
                        = & \ddTheta \sum_{h^{1:K}} Q(h^{1:K}|x) \Lhath  \nonumber \\
                        = & \sum_{h^{1:K}} \Big [ \Lhath  \ddTheta Q(h^{1:K}|x) + \nonumber \\
                          & Q(h^{1:K}|x)  \ddTheta \Lhath \Big].
\end{align}
Using the identity $\ddTheta g(x) = g(x) \ddTheta \log g(x)$, we can express
the gradient of $Q(h^{1:K}|x)$ as
\begin{align}
    \label{eqn:gradQ}
    \ddTheta Q(h^{1:K}|x) = & Q(h^{1:K}|x) \ddTheta \log Q(h^{1:K}|x) \nonumber \\
                          = & Q(h^{1:K}|x) \ddTheta \log \prod_{j=1}^{K} Q(h^j|x) \nonumber \\
                          = & Q(h^{1:K}|x) \sum_{j=1}^{K} \ddTheta \log Q(h^j|x).
\end{align}
We use the chain rule along with the same identity to compute the gradient of $\Lhath$:
\begin{align}
    \label{eqn:gradL}
    \ddTheta \Lhath = & \ddTheta \log \frac{1}{K} \sum_{j=1}^{K} f(x, h^j) \nonumber \\
                    = & \frac{1}{\sum_{i=1}^{K} f(x, h^i)} \sum_{j=1}^{K} \ddTheta f(x, h^j) \nonumber\\
                    = & \frac{1}{\sum_{i=1}^{K} f(x, h^i)} \sum_{j=1}^{K} f(x, h^j) \ddTheta \log f(x, h^j) \nonumber\\
                    = & \sum_{j=1}^{K} \Wtil^j \ddTheta \log f(x, h^j)
\end{align}
where  $\Wtil^j \equiv \frac{f(x, h^j)}{\sum_{i=1}^K f(x, h^i)}$.
Substituting \eqn{eqn:gradQ} and \eqn{eqn:gradL} into \eqn{eqn:gradVarL} we obtain
\begin{align}
    \ddTheta \varL^K(x) = & \sum_{h^{1:K}} \Big ( \Lhath  Q(h^{1:K}|x) \sum_{j=1}^{K} \ddTheta \log Q(h^j|x) + \nonumber \\
                          & Q(h^{1:K}|x)  \sum_{j=1}^{K} \Wtil^j \ddTheta \log f(x, h^j) \Big), \nonumber \\
                        = & \sum_{h^{1:K}} Q(h^{1:K}|x) \Lhath  \sum_{j=1}^{K} \ddTheta \log Q(h^j|x) + \nonumber \\
                          & \sum_{h^{1:K}} Q(h^{1:K}|x) \sum_{j=1}^{K} \Wtil^j \ddTheta \log f(x, h^j), \nonumber \\
                        = & \EQ \left[ \sum_j \Lhath \ddTheta \log Q(h^j|x) \right] +  \nonumber \\
                          & \EQ \left[ \sum_j \Wtil^j \ddTheta \log f(x, h^j) \right].
\end{align}

\section*{E. Structured output prediction: digit completions}

\begin{figure*}
\includegraphics[width=.99\textwidth]{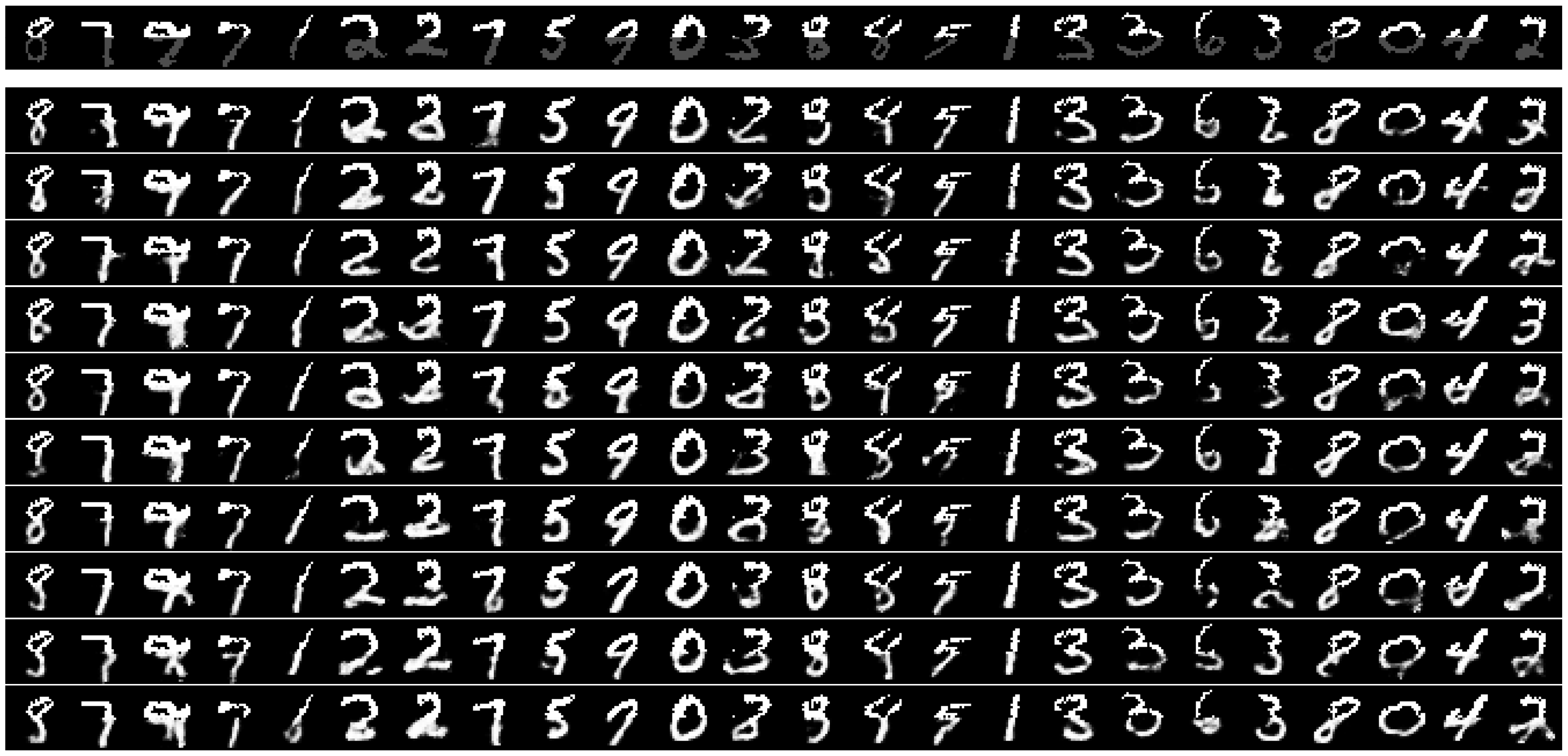}
\caption{Structured output prediction: Conditional completions generated by
sampling from a three-layer SBN trained using VIMCO with the 20-sample objective.
The top row shows the original full digit images. The remaining rows combine
the top half from the original image with the bottom half generated from the model.
}
\label{fig:mnist_completions}
\end{figure*}

\fig{fig:mnist_completions} shows multiple completions for the same set of top
digit image halves generated using a three-layer (200-200-200) SBN trained using VIMCO 
with the 20-sample objective. The completions were obtained by computing
observation probabilities based on a single sample from the prior. The
variability of the completions shows how the model captured the multimodality
of the data distribution.

%\renewcommand{\bibsection}{\section*{References}}
%\bibliography{nvilk}
%\bibliographystyle{icml2016}